%% file: paper.tex
\documentclass[runningheads]{llncs}
\usepackage{graphicx}
\usepackage{amsmath,amssymb} 
\usepackage{color}
\usepackage{subfigure}
\usepackage{caption}
\usepackage[para]{threeparttable} 

\usepackage[utf8x]{inputenc}
\PrerenderUnicode{é}

\usepackage[pagebackref=true,breaklinks=true,letterpaper=true,colorlinks,bookmarks=false]{hyperref}
\usepackage{csquotes}

\graphicspath{{detections/}}

\usepackage{soul}

\usepackage{xspace}

\makeatletter
\DeclareRobustCommand\onedot{\futurelet\@let@token\@onedot}
\def\@onedot{\ifx\@let@token.\else.\null\fi\xspace}

\def\eg{\emph{e.g}\onedot} 
\def\ie{\emph{i.e}\onedot}

\def\wrt{w.r.t\onedot} 
\def\etal{\emph{et al}\onedot}
\makeatother

\DeclareMathOperator{\atan2}{atan2}

\newcommand{\norm}[1]{\left\lVert#1\right\rVert}

\def\deg{$^\circ\ $}

\begin{document}
\pagestyle{headings}
\mainmatter

\title{Eliminating the Blind Spot: Adapting 3D Object Detection and Monocular Depth Estimation to 360\deg Panoramic Imagery} 

\titlerunning{Adapting 3D Object Detection and Depth Estimation to Panoramic Imagery}


\author{Grégoire Payen de La Garanderie,
Amir Atapour Abarghouei, \\ and
Toby P. Breckon}
\authorrunning{G. Payen de La Garanderie, A. Atapour Abarghouei, T. Breckon}

\institute{Department of Computer Science \\
    Durham University \\
    \email{ gregoire.p.payen-de-la-garander@durham.ac.uk \\ \{amir.atapour-abarghouei,toby.breckon\}@durham.ac.uk}
}

\maketitle

\begin{abstract}
Recent automotive vision work has focused almost exclusively on processing forward-facing cameras. However, future autonomous vehicles will not be viable without a more comprehensive surround sensing, akin to a human driver, as can be provided by 360\deg panoramic cameras.  We present an approach to adapt contemporary deep network architectures developed on conventional rectilinear imagery to work on equirectangular 360\deg panoramic imagery. To address the lack of annotated panoramic automotive datasets availability, we adapt a contemporary automotive dataset, via style and projection transformations, to facilitate the cross-domain retraining of contemporary algorithms for panoramic imagery. Following this approach we retrain and adapt existing architectures to recover scene depth and 3D pose of vehicles from monocular panoramic imagery without any panoramic training labels or calibration parameters. Our approach is evaluated qualitatively on crowd-sourced panoramic images and quantitatively using an automotive environment simulator to provide the first benchmark for such techniques within panoramic imagery.

\end{abstract}

\begin{figure}
    \includegraphics[width=\textwidth]{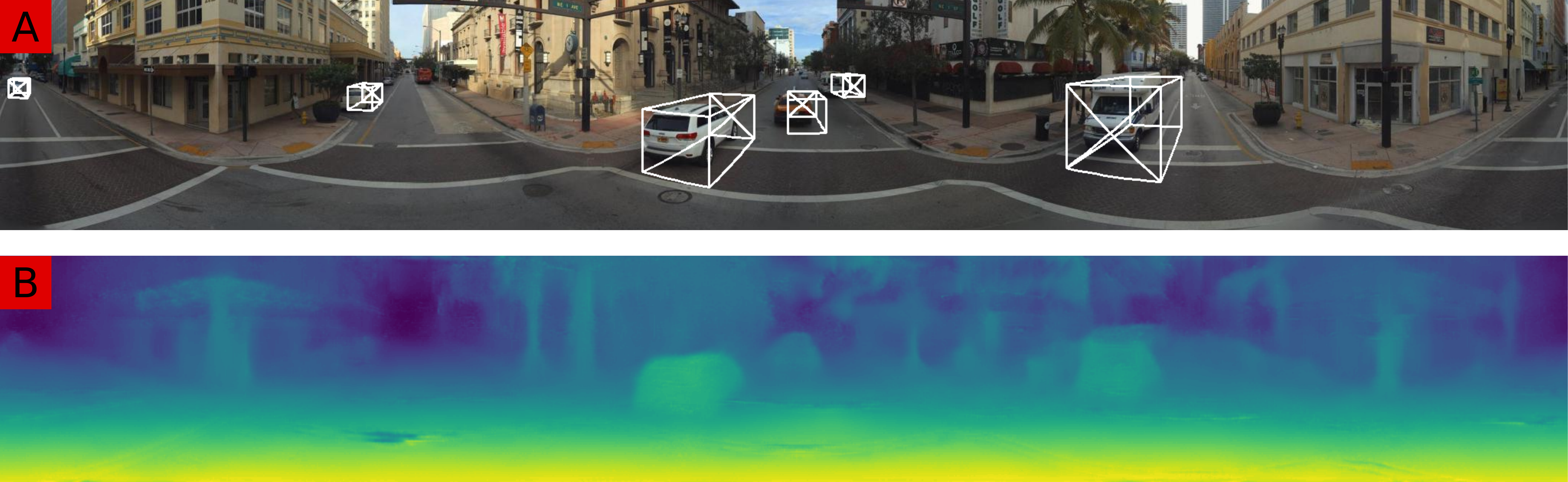}
    \caption{\label{fig:firstexample} Our monocular panoramic image approach.
    A: 3D object detection. B: depth recovery. }
\end{figure}

\section{Introduction}

Recent automotive computer vision work (object detection \cite{RenFasterRCNNRealTime2015,RenAccurateSingleStage2017}, segmentation \cite{BadrinarayananSegNetDeepConvolutional2015}, stereo vision \cite{KendallEndtoEndLearningGeometry2017,PangCascadeResidualLearning2017}, monocular depth estimation \cite{LadickyPullingThingsout2014,GodardUnsupervisedMonocularDepth2017,Atapour-AbarghoueiRealTimeMonocularDepth2018}) has focused almost exclusively on the processing of forward-facing rectified rectilinear vehicle mounted cameras. Indeed by sharp contrast to the abundance of common evaluation criteria and datasets for the forward-facing camera case \cite{GeigerAreweready2012,GeigerVisionmeetsrobotics2013,BrostowSemanticobjectclasses2009,KondermannHCIBenchmarkSuite2016,NeuholdMapillaryVistasDataset2017,FisherBerkeleyDataDrive2018,AustrianInstituteofTechnologyWildDashBenchmark2018}, there are no annotated evaluation datasets or frameworks for any of these tasks using 360\deg view panoramic cameras.

However, varying levels of future vehicle autonomy will require full 360\deg situa\-tional awareness, akin to that of the human driver of today, in order to be able to function across complex and challenging driving environments. One popularly conceived idea of capturing this awareness is to use active sensing in the form of 360\deg LIDAR, however this is currently an expensive, low-resolution method which does not encompass the richness of visual information required for high fidelity semantic scene understanding. An alternative is to fuse the information from multiple cameras surrounding the vehicle \cite{HamiltonGeneralizeddynamicobject2016} and such methods have been used to fuse between a forward-facing camera and LIDAR \cite{ChenMultiView3DObject2017,GonzalezOnBoardObjectDetection2017}. However, here opportunities are lost to share visual information in early stages of the pipeline with further computational redundancy due to overlapping fields of view. Alternatively the imagery from a multiview setup can be stitched into a 360\deg panorama \cite{BrownMultiimagematchingusing2005}. A roof mounted on-vehicle panoramic camera offers superior angular resolution compared to any LIDAR, is 1--2 orders of magnitude lower cost and provides rich scene colour and texture information that enables full semantic scene understanding \cite{JanaiComputerVisionAutonomous2017}.

Panoramic images are typically represented using an equirectangular projection (Fig. \ref{fig:firstexample}A); in contrast, a conventional camera uses a rectilinear projection. In this projection, the image-space coordinates are proportional to latitude and longitude of observed points rather than the usual projection onto a focal plane  as shown in Fig. \ref{fig:firstexample}A.

Recent work on panoramic images has largely focused on indoor scene understanding \cite{ZhangPanoContextWholeRoom3D2014,XuPano2CADRoomLayout2017}, panoramic to rectilinear video conversion \cite{SuPano2VidAutomaticCinematography2016,HuDeep360Pilot2017,LaiSemanticdrivenGenerationHyperlapse2018} and dual camera 360\deg stereo depth recovery \cite{HaneRealTimeDirectDense2014,MatzenLowcost360Stereo2017}. However, no work to date has explicitly tackled contemporary automotive sensing problems.

By contrast, we present an approach to adapt existing deep architectures, such as convolutional neural networks (CNN) \cite{CaiUnifiedMultiscaleDeep2016,GodardUnsupervisedMonocularDepth2017}, developed on rectilinear imagery to operate on equirectangular panoramic imagery. Due to the lack of explicit annotated panoramic automotive training datasets, we show how to reuse existing non-panoramic datasets such as KITTI \cite{GeigerAreweready2012,GeigerVisionmeetsrobotics2013} using style and projection transformations, to facilitate the cross-domain retraining of contemporary algorithms for panoramic imagery. We apply this technique to estimate dense monocular depth (see example in Fig. \ref{fig:firstexample}B) and to recover the full 3D pose of vehicles (Fig. \ref{fig:firstexample}B) from panoramic imagery. Additionally, our work provides the first performance benchmark for the use of these techniques on 360\deg panoramic imagery acting as a key driver for future research on this topic. Our technique is evaluated qualitatively on crowd-sourced 360\deg panoramic images from Mapillary \cite{MapillaryMapillaryResearch} and quantitatively using ground truth from the CARLA \cite{DosovitskiyCARLAOpenUrban2017} high fidelity automotive environment simulator\footnote{for future comparison our code, models and evaluation data is publicly available at: \url{https://gdlg.github.io/panoramic}}.

\section{Related Work}

Related work is considered within panoramic imagery (Section \ref{sec:relwork:panoramic_imagery}), monocular 3D object detection (Section \ref{sec:relwork:3d_object_detection}),
monocular depth recovery (Section \ref{sec:relwork:mono_depth_estimation}) and domain adaptation (Section \ref{sec:relwork:style_transfer}).

\subsection{Object Detection within Panoramic Imagery}\label{sec:relwork:panoramic_imagery}

Even though significant strides have been made in rectilinear image object proposal \cite{hosang2016makes} and object detection methods utilizing deep networks \cite{RenFasterRCNNRealTime2015,sermanet2013overfeat,girshick2014rich,girshick2015fast,he2017mask,CaiUnifiedMultiscaleDeep2016}, comparatively limited literature exists within panoramic imagery.

Deng \etal \cite{DengObjectdetectionpanoramic2017} adapted, trained and evaluated Faster R-CNN \cite{RenFasterRCNNRealTime2015} on a new dataset of 2,000 indoor panoramic images for 2D object detection. However their approach does not handle the special case of object  wrap-around at the equirectangular image boundaries.  

Recently, object detection and segmentation has been applied directly to equirectangular panoramic images to provide object detection and saliency in the context of virtual cinematography \cite{HuDeep360Pilot2017,LaiSemanticdrivenGenerationHyperlapse2018} using pre-trained detectors such as Faster R-CNN \cite{RenFasterRCNNRealTime2015}. Su and Grauman \cite{SuFlat2SphereLearningSpherical2017} introduce a Flat2Sphere technique to train a spherical CNN to imitate the results of an existing CNN facilitating large object detection at any angle.

In contemporary automotive sensing problems, the required vertical field of view is small as neither the view above the horizon nor the view directly underneath the camera have any useful information for those problems. Therefore, the additional complexity of the spherical CNN introduced by \cite{SuFlat2SphereLearningSpherical2017} is not needed in the specific automotive context. Instead we show how to reuse existing deep architectures without requiring any significant architectural changes.

\subsection{Monocular 3D Object Detection} \label{sec:relwork:3d_object_detection}

Prior work on 3D pose regression in panorama is mostly focused on indoor scene reconstruction such as PanoContext by Zhang \etal \cite{ZhangPanoContextWholeRoom3D2014} and Pano2CAD by Xu \etal \cite{XuPano2CADRoomLayout2017}. The latter retrieves the object poses by regression using a bank of known CAD (Computer-Aided Design) models. In contrast, our method does not require any \textit{a priori} knowledge of the object geometry.

Contemporary end-to-end CNN driven detection approaches are based on the R-CNN architecture introduced by Girshick \cite{GirshickRichFeatureHierarchies2014}. Successive improvements from Fast-RCNN \cite{GirshickFastRCNN2015} and Faster-RCNN \cite{RenFasterRCNNRealTime2015} increased the performance by respectively sharing feature maps across proposals and generating the proposals using a Region Proposal Network (RPN) instead of traditional techniques based on sliding windows. This allowed a unified end-to-end training of the network to solve the combined detection and classification tasks. More recently, Yang \etal \cite{YangExploitalllayers2016} and Cai \etal \cite{CaiUnifiedMultiscaleDeep2016} introduced a multi-scale approach by pooling the region proposals from multiple layers in order to reduce the number of proposals needed as well as to improve performance on smaller objects such as distant objects.

While most of the work has focused on 2D detection, the work of Chen \etal \cite{Chen3DObjectProposals2015,ChenMultiView3DObject2017} leverages 3D pointcloud information gained either from stereo or LIDAR modalities to generate 3D proposals which are pruned using Fast R-CNN. Whereas these works use complex arrangements using stereo vision, handcrafted features or 3D model regression, recent advances \cite{ChenMonocular3DObject2016,Mousavian3DBoundingBox2016,ChabotDeepMANTACoarsetofine2017} show that it is actually possible to recover the 3D pose from monocular imagery. Chen \etal \cite{ChenMonocular3DObject2016} use post-processing of the proposals within an energy minimization framework assuming that the ground plane is known. Chabot \etal \cite{ChabotDeepMANTACoarsetofine2017} use 3D CAD models as templates to regress the 3D pose of an object given part detections; while Mousavian \etal \cite{Mousavian3DBoundingBox2016} show the 3D pose can be recovered without any template assumptions using carefully-expressed geometrical constraints.
In this work, we propose a new approach, similar to \cite{Mousavian3DBoundingBox2016}, however without explicitly-expressed geometrical constraints, which performs on both rectilinear and equirectangular panoramic imagery without any knowledge of the ground plane position with respect to the camera.

\subsection{Monocular Depth Estimation} \label{sec:relwork:mono_depth_estimation}

Traditionally dense scene depth is recovered using multi-view approaches such as structure-from-motion and stereo vision \cite{Scharsteintaxonomyevaluationdense2002}, relying on an explicit handling of geometrical constraints between multiple calibrated views. However recently with the advance of deep learning, it has been shown that dense scene depth can also be recovered from monocular imagery.

After the initial success of classical learning-based techniques such as \cite{saxena2006learning,saxena2009make3d}, depth recovery was first approached as a supervised learning problem by the depth classifier of Ladick\'y \etal \cite{LadickyPullingThingsout2014} and deep learning-based approaches such as \cite{EigenDepthMapPrediction2014,LiuLearningDepthSingle2016}. However, these techniques are based on the availability of high-quality ground truth depth maps, which are difficult to obtain. In order to combat the ground truth data issue, the method in \cite{Atapour-AbarghoueiRealTimeMonocularDepth2018} relies on readily-available high-resolution synthetic depth maps captured from a virtual environment and domain transfer to resolve the problem of domain bias.

On the other hand, other monocular depth estimation methods have recently emerged that are capable of performing depth recovery without the need for large quantities of ground truth depth data. Zhou \etal \cite{zhou2017unsupervised} estimate monocular depth and ego-motion using depth and pose prediction networks that are trained via view synthesis. The approach proposed in \cite{kuznietsov2017semi} utilizes a deep network semi-supervised by sparse ground truth depth and then reinforced within a stereo framework to recover dense depth information.

Godard \etal \cite{GodardUnsupervisedMonocularDepth2017} train their model based on left-right consistency inside a stereo image pair during training. At inference time, however, the model solely relies on a single monocular image to estimate a dense depth map. Even though said approach is primarily designed to deal with rectilinear images, in this work we further adapt this model to perform depth estimation on equirectangular panoramic images.

\subsection{Domain Adaptation and Style Transfer} \label{sec:relwork:style_transfer} \label{sec:style_transfer}

Machine learning architectures trained on one dataset do not necessarily transfer well to a new dataset -- a problem known as dataset bias \cite{TorralbaUnbiasedlookdataset2011} or covariate shift \cite{GrettonCovariateShiftKernel2008}. A simple solution to dataset bias would be fine-tuning the trained model using the new data but that often requires large quantities of ground truth, which are not always readily-available.

While many strategies have been proposed to reduce the feature distributions between the two data domains \cite{LongLearningTransferableFeatures2015,GhifaryDeepReconstructionClassificationNetworks2016,DonahueAdversarialFeatureLearning2016,TzengAdversarialDiscriminativeDomain2017}, a novel solution was recently proposed in \cite{Atapour-AbarghoueiRealTimeMonocularDepth2018} that uses image style transfer as a means to circumvent the data domain bias.

Image style transfer was first proposed by Gatys \etal \cite{GatysImageStyleTransfer2016} but since then remarkable advances have been made in the field \cite{JohnsonPerceptualLossesRealTime2016,UlyanovTextureNetworksFeedforward2016,DumoulinLearnedRepresentationArtistic2016,GhiasiExploringstructurerealtime2017}. In this work, we attempt to transform existing rectilinear training images (such as KITTI \cite{GeigerAreweready2012,GeigerVisionmeetsrobotics2013}) to share the same style as our panoramic destination domain (Mapillary \cite{MapillaryMapillaryResearch}). However, these two datasets have been captured in different places and share no registration relationship. As demonstrated in \cite{Atapour-AbarghoueiRealTimeMonocularDepth2018,HoffmanCyCADACycleConsistentAdversarial2017}, unpaired image style transfer solved by CycleGAN \cite{ZhuUnpairedImagetoImageTranslationUnpairedImage-to-ImageTranslationusingCycle-ConsistentAdversarial}, can be used to transfer the style between two data domains that possess approximately similar content.

\subsection{Proposed Contributions}

Overall the main contributions, against the state of the art \cite{CaiUnifiedMultiscaleDeep2016,GodardUnsupervisedMonocularDepth2017,GeigerAreweready2012,GeigerVisionmeetsrobotics2013,Mousavian3DBoundingBox2016,GodardUnsupervisedMonocularDepth2017,DosovitskiyCARLAOpenUrban2017}, presented in this work are:
\begin{itemize}
    \item a novel approach to convert deep network architectures \cite{CaiUnifiedMultiscaleDeep2016,GodardUnsupervisedMonocularDepth2017} operating on rectilinear images for equirectangular panoramic images based on style and projection transformations;
    \item a novel approach to reuse and adapt existing datasets \cite{GeigerAreweready2012,GeigerVisionmeetsrobotics2013} in order to train models for panoramic imagery;
    \item the subsequent application of these approaches for monocular 3D object detection using a simpler formulation than earlier work \cite{Mousavian3DBoundingBox2016}, additionally operable on conventional imagery without modification;
    \item further application of these techniques to monocular depth recovery using an adaptation of the rectilinear imagery approach of Godard \etal \cite{GodardUnsupervisedMonocularDepth2017};
    \item provision of the first performance benchmark based on a new synthetic evaluation dataset (based on CARLA \cite{DosovitskiyCARLAOpenUrban2017}) for this new challenging task of automotive panoramic imagery depth recovery and object detection evaluation.
\end{itemize}

\section{Approach}

We first describe the mathematical projections underlining rectilinear and equirectangular projections and the relationship between the two required to enable our approach within panoramic imagery (Sec. \ref{sec:projections}). Subsequently we describe the dataset adaptation (Sec. \ref{sec:dataset_adaptation}), its application to monocular 3D pose recovery (Sec. \ref{sec:3d_object_detection}) and depth estimation (Sec. \ref{sec:monocular_depth_recovery}) and finally the architectural modifications required for inference within panoramic imagery (Sec. \ref{sec:network_adaptation}).

\subsection{Rectilinear and Equirectangular Projections}\label{sec:projections}
\def\ulin{u_{lin}}
\def\vlin{v_{lin}}
\def\Tequi{T_{equi}}
\def\uequi{u_{equi}}
\def\vequi{v_{equi}}

Projection using a classical rectified rectilinear camera is typically defined in terms of its camera matrix $P$. Given the Cartesian coordinates $(x,y,z)$ of a 3D scene point in camera space, its projection $(\ulin,\vlin)$ is defined as:

\begin{equation}
    \label{eq:usual_proj}
    \begin{bmatrix} \ulin \\ \vlin \end{bmatrix} = \left\lfloor P \cdot \begin{bmatrix}x \\ y \\ z\end{bmatrix} \right\rfloor
\end{equation}

where $\lfloor\cdot\rfloor$ denotes the homogeneous normalization of the vector by its last component.
The camera matrix $P$ is conventionally defined as:

\begin{equation}
    P = \begin{bmatrix}
        f & 0 & c_x \\
        0 & f & c_y \\
        0 & 0 & 1
    \end{bmatrix}
\end{equation}

where $f$ and $(c_x,c_y)$ are respectively the focal length and the principal point of the camera.

The rectilinear projection as defined in Eqn. \ref{eq:usual_proj} is advantageous because the camera matrix $P$ can be combined with further image and object space transformations into a single linear transformation followed by an homogeneous normalization. However, this transformation can also be written as:

\begin{equation}
    \label{eq:lin_proj}
    \begin{bmatrix} \ulin \\ \vlin \end{bmatrix} = P \cdot \begin{bmatrix} x/z \\ y/z \\ 1\end{bmatrix}
\end{equation}

This formulation (Eqn. \ref{eq:lin_proj}) is convenient because the image-space coordinates are expressed in terms of the ratio $x/z$ and $y/z$ which are the same regardless of the distance from the 3D scene point to the camera.

In contrast, the equirectangular projection is defined in terms of the longitude and latitude of the point. The longitude and latitude, respectively $(\lambda,\phi)$, are defined as:
\begin{align}
    \lambda & = \text{arctan}\ x/z \\
    \label{eq:usual_phi} \phi    & = \text{arcsin}\ y/r \quad \text{where } r = (x^2 + y^2 + z^2)^{\frac{1}{2}}
\end{align}

The latitude definition in Eqn. \ref{eq:usual_phi} can be conveniently rewritten in terms of the ratios $x/z$ and $y/z$ as in Eqn. \ref{eq:lin_proj} for rectilinear projections:
\begin{equation}
    \phi    = \text{arcsin}\ \frac{y/z}{r} \quad  \text{where } r = (x/z^2 + y/z^2 + 1^2)^{\frac{1}{2}}
\end{equation}

For the sake of simplicity, this computation of the latitude and longitude from the Cartesian coordinates can be represented as a function $\Gamma$:

\begin{equation}
    \begin{bmatrix} \lambda \\ \phi \end{bmatrix} = \Gamma\left(\begin{bmatrix}x \\ y \\ z \end{bmatrix}\right) = \Gamma\left(\begin{bmatrix}x/z \\ y/z \\ 1 \end{bmatrix}\right)
\end{equation}

Finally, we define an image transformation matrix $\Tequi$ which transforms the longitude and latitude to image space coordinates $(\uequi, \vequi)$:
\begin{equation}
    \label{eq:equi_proj}
    \begin{bmatrix} \uequi \\ \vequi \\ 1\end{bmatrix} = \Tequi \cdot \begin{bmatrix} \lambda \\ \phi \\ 1 \end{bmatrix} = \Tequi \cdot \Gamma\left(\begin{bmatrix}x/z \\ y/z \\ 1 \end{bmatrix}\right)
\end{equation}

The matrix $\Tequi$ can be defined as:
\begin{equation}
    \Tequi =
    \begin{bmatrix}
        \alpha & 0 & c_\lambda \\
        0 & \alpha & c_\phi \\
        0 &      0 & 1
    \end{bmatrix}
\end{equation}

where $\alpha$ is an angular resolution parameter akin to the focal length. Like the focal length, it can be defined in terms of the field of view:
\begin{equation}
    \alpha = \text{fov}_\lambda / w = \text{fov}_\phi / h
\end{equation}

where $\text{fov}_\lambda, \text{fov}_\phi, w, h$ are respectively the image horizontal field of view, vertical field of view; width and height.
In contrast to rectilinear imagery, where the focal length is difficult to determine without any kind of camera calibration, the equirectangular imagery, commonly generated by panoramic cameras from the raw dual-fisheye pair, can be readily used without any prior calibration because the angular resolution $\alpha = 2\pi/w$ depends only on the image width.
Therefore, approaches that would require some knowledge of the camera intrinsics of rectilinear images (\eg monocular depth estimation) can be readily used on any 360\deg panoramic image without any prior calibration.

By coupling the definitions of both the rectilinear and equirectangular projections in terms of the ratios $x/z$ and $y/z$ (Eqn. \ref{eq:lin_proj} \& \ref{eq:equi_proj}), we establish the relationship between the coordinates in the rectilinear projection and equirectangular projection for the given matrices $P$ and $\Tequi$:
\begin{equation}
    \begin{bmatrix} \uequi \\ \vequi \\ 1\end{bmatrix} = \Tequi \cdot \Gamma\left(P^{-1} \cdot \begin{bmatrix}\ulin \\ \vlin \\ 1 \end{bmatrix}\right)
\end{equation}

This enables us to reproject an image from one projection to another, such as from the rectilinear image (Fig. \ref{fig:overallprocess}A) to an equirectangular image (Fig. \ref{fig:overallprocess}C) and vice versa --- a key enabler for the application of our approach within panoramic imagery.

\subsection{Dataset Adaptation}\label{sec:dataset_adaptation}
\begin{figure}[b]
    \centering
    \includegraphics[width=\textwidth]{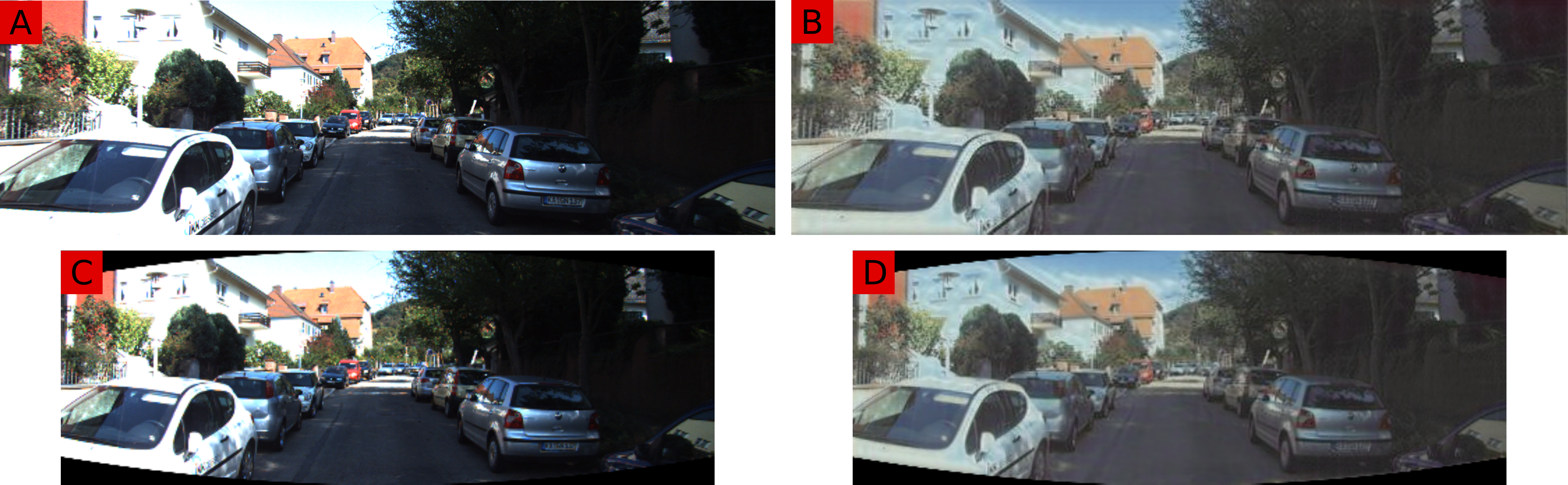}

    \caption{\label{fig:overallprocess}Output of each step of the adaptation of an image from the KITTI dataset:
        {\small A: No tranformation, B: Style transfer, C: Projection transfer, D: Style and projection}}
\end{figure}

In our approach, the source domain is the KITTI \cite{GeigerAreweready2012,GeigerVisionmeetsrobotics2013} dataset of rectilinear images captured using a front-facing camera rig (1242$\times$375 image resolution; 82.5\deg horizontal FoV and 29.7\deg vertical FoV);
while our target domain consist of 30,000 images from the Mapillary \cite{MapillaryMapillaryResearch} crowd-sourced street-level imagery (2048$\times$300 image resolution; 360\deg $\times$ 52.7\deg FoV). These latter images are cropped vertically from 180\deg down to 52.7\deg which is more suitable for automotive problems. This reduced panorama has an angular coverage 7.7 times larger than our source KITTI imagery.
Due to the lack of annotated labels for our target domain, we adapt the source domain dataset to train deep architectures for panoramic imagery via a methodology based on projection and style transformations.

Due to dataset bias \cite{TorralbaUnbiasedlookdataset2011}, training on the original source domain is unlikely to perform well on the target domain. Furthermore our target is relatively low resolution and has numerous compression artefacts not present in the source domain -- present due to the practicality of 360\deg image transmission and storage.
To improve generalization to the target domain, we transform the source domain to look similar to imagery from our target domain via a two-step process.

The first step transfers the style of our target domain (reprojected as rectilinear images) onto each image from the source domain (Fig. \ref{fig:overallprocess}A); resulting images are shown in Fig. \ref{fig:overallprocess}B. We use the work of Zhu \etal on CycleGAN \cite{ZhuUnpairedImagetoImageTranslationUnpairedImage-to-ImageTranslationusingCycle-ConsistentAdversarial} to learn a transformation back and forth between the two unpaired domains. Subsequently, this transformation model is used to transfer the style of our target domain onto all the images from our source domain. 
In essence, the style transfer introduces a tone mapping and imitates compression artifacts present in most panoramic images while preserving the actual geometry. Without the use of style transfer, the weights are biased toward high-quality imagery and perform poorly on low-quality images.

The second step reprojects the style-transferred images (Fig. \ref{fig:overallprocess}B) and annotations from the source domain rectilinear projection to an equirectangular projection (Fig. \ref{fig:overallprocess}D). The transformed images represent small subregions (FoV: 82.5\deg$\times$29.7\deg) of a larger panorama.
While this set of transformed images does not cover the full panorama, we find that they are sufficient to train deep architectures that perform well on full size panoramic imagery.

\subsection{3D Object Detection} \label{sec:3d_object_detection}

For 3D detection, we use a network by Cai \etal \cite{CaiUnifiedMultiscaleDeep2016} based on Faster R-CNN \cite{RenFasterRCNNRealTime2015}. This network generates a sequence of detection proposals using a Region Proposal Network (RPN) and then pools a subregion around each proposal to further regress the proposal 2D location.
We extend this network to support 3D object pose regression. Uniquely, our extended network can be used on either rectilinear or equirectangular imagery without any changes to the network itself, instead only requiring a change to the interpretation of the output for subsequent rectilinear or equirectangular imagery use.

While Mousavian \etal \cite{Mousavian3DBoundingBox2016} shows that 3D pose can be estimated without any assumptions of known 3D templates, their algorithm relies on geometrical properties. In contrast, we regress the 3D pose directly, simplifying the computation and making it easier to adapt to equirectangular images.

Here, we directly regress the 3D dimensions (width, length and height) in meters of each object using a fully-connected layer as well as the orientation as per \cite{Mousavian3DBoundingBox2016}. Moreover, instead of relying on geometrical assumptions, we also regress the object disparity $d_{lin} = \frac r f$ which is the inverse of the distance $r$ multiplied by the focal length $f$. For equirectangular imagery, we use a similar definition $d_{equi} = \frac r \alpha$ substituting the angular resolution for the focal length. Using a fully-connected layer connected to the last common layer defined in \cite{CaiUnifiedMultiscaleDeep2016}, we learn coefficients $a,b$ such that the disparity $d$ can be expressed as:
\begin{equation}
d = a h_{roi} + b
\end{equation}
where $h_{roi}$ is the height of the region proposal generated by the RPN. To simplify the computation, we also learn the 2D projection of the centre of the object onto the image $(u,v)$ using another fully-connected layer. As a result, we can recover the actual 3D position $(x,y,z)$ using:

\begin{align}
    [x,y,z]^T & = r \cdot u[P^{-1} \cdot [\ulin,\vlin,1]^T] & \text{ \emph{rectilinear case}} \\
    [x,y,z]^T & = r \cdot u[\Gamma^{-1} (\Tequi^{-1} \cdot [\uequi,\vequi,1]^T)] & \text{ \emph{equirectangular case}}
\end{align}
where $u[v] = \frac{v}{\norm{v}}$ is the unit vector in the direction of $v$.

For network training of our model, we additionally use data augmentation including image cropping and resizing as defined by \cite{CaiUnifiedMultiscaleDeep2016}. Any of those operations on the image must be accompanied by the corresponding transformation of the corresponding camera matrix $P$ or $\Tequi$ in order to facilitate effective training.

As noted by Mousavian \etal \cite{Mousavian3DBoundingBox2016}, distant objects (far) pose a significant challenge for reliable detection of the absolute orientation (\ie relative front to back directional pose). Confronted with such an ambiguity (absolute directional orientation), a naive regression using the mean-square error would choose the average of the two extrema rather than the most likely extremum. To circumvent this problem, given the object yaw $\theta$ (orientation on the ground plane), we instead learn $c = \cos^2\theta$ an $s = \sin^2\theta$ which are both independent of the directionality. Noting that $\cos^2 \theta + \sin^2 \theta$ = 1, $c$ and $s$ can be very conveniently learned with a fully-connected layer followed by a \emph{Softmax()} layer. For each pair $(s,c)$, there are four possible angles each in a different quadrant depending on the sign of the sine and cosine:

\begin{equation}\label{equ:yaw}
\hat\theta = \atan2(\pm \sqrt s, \pm \sqrt c)
\end{equation}

We further discriminate between the four quadrants:
\begin{equation}
    \{(-1,-1),(-1,1),(1,-1),(1,1) \}
\end{equation} using a separate classifier consisting of a fully-connected layer followed by a \emph{Softmax()} classification layer.

Our entire network, comprising the architecture of \cite{CaiUnifiedMultiscaleDeep2016} and our 3D pose regression extension, is fine-tuned end-to-end using a multi-task loss over 6 sets of heterogeneous network outputs: \emph{class} and \emph{quadrant classification} are learned via cross entropy loss while \emph{bounding-box position}, \emph{object centre}, \emph{distance}, \emph{orientation} are dependent on a mean-square loss. As a result, it would be time-consuming to manually tune the multi-task loss weights, therefore we use the methodology of \cite{KendallMultiTaskLearningUsing2017} to dynamically adjust the multi-task weights during training based on homoscedastic uncertainty without any use of manual hyperparameters.

\subsection{Monocular Depth Recovery}\label{sec:monocular_depth_recovery}

We rely on the approach of Godard \etal \cite{GodardUnsupervisedMonocularDepth2017} which was originally trained and tested on the rectilinear stereo imagery of the KITTI dataset \cite{GeigerAreweready2012}. We reuse the same architecture and retrain it on our domain-adapted KITTI dataset constructed using the methodology of Sec. \ref{sec:dataset_adaptation}.

Following the original work \cite{GodardUnsupervisedMonocularDepth2017}, the loss function is based on a left-right consistency check between a pair of stereo images. In our new dataset, both stereo images have been warped to an equirectangular projection as well as depth smoothness constraints.
While Godard \etal uses the stereo disparity $d_{stereo} = \frac{fB}{zw}$ where $f$ is the focal length, $B$ the stereo baseline and $w$ the width of the image, we replace the focal length with the angular resolution: $d_{equi} = \frac{\alpha B}{rw}$.

Given a point $p_l=(u_l,v_l)^T$, the corresponding point $p_r=(u_r,v_r)^T$ for a given disparity $d$ can be calculated as:
\begin{equation}\label{equ:equirect_stereo}
    p_r = \Tequi \cdot \Gamma\left[ u\left[\Gamma^{-1} (\Tequi^{-1} \cdot p_l)\right] + \left[\frac{d_{equi}w}{\alpha},0,0\right]^T \right]
\end{equation}

with definitions as per Sec. \ref{sec:projections}. The corresponding point $p_r$ in Eqn. \ref{equ:equirect_stereo} is differentiable \wrt $d_{equi}$ and is used for the left/right consistency check instead of the original formulation presented in \cite{GodardUnsupervisedMonocularDepth2017}. This alternative formulation (Eqn. \ref{sec:projections}) explicitly takes into account that the epipolar lines in a conventional rectilinear stereo setup are transformed to epipolar curves within panoramic imagery, hence enabling the adaptation of monocular depth prediction \cite{GodardUnsupervisedMonocularDepth2017} to this case.

\subsection{360\deg Network Adaptation} \label{sec:network_adaptation}

\begin{figure}[b]
    \centering
    \subfigure[\label{fig:360folding}  A 360\deg equirectangular image can be folded over itself until the ends meet.]{
            \includegraphics[height=2.5cm]{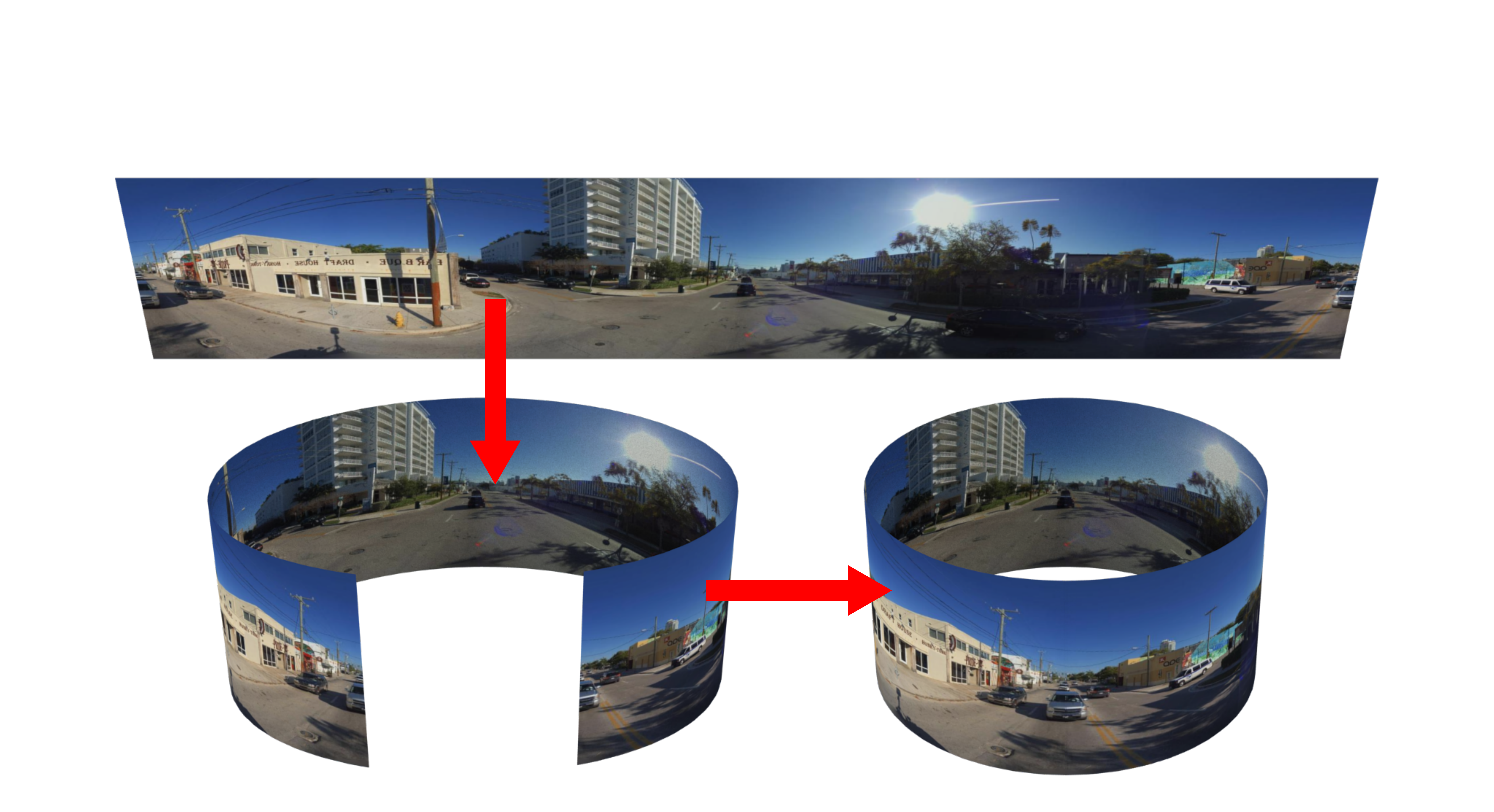}
        }\hspace{1cm}
    \subfigure[\label{fig:kernel3x3}A $3\times 3$ convolution kernel, a column of padding copied from the other side is added at each extremity]{\centering
            \includegraphics[height=2.5cm]{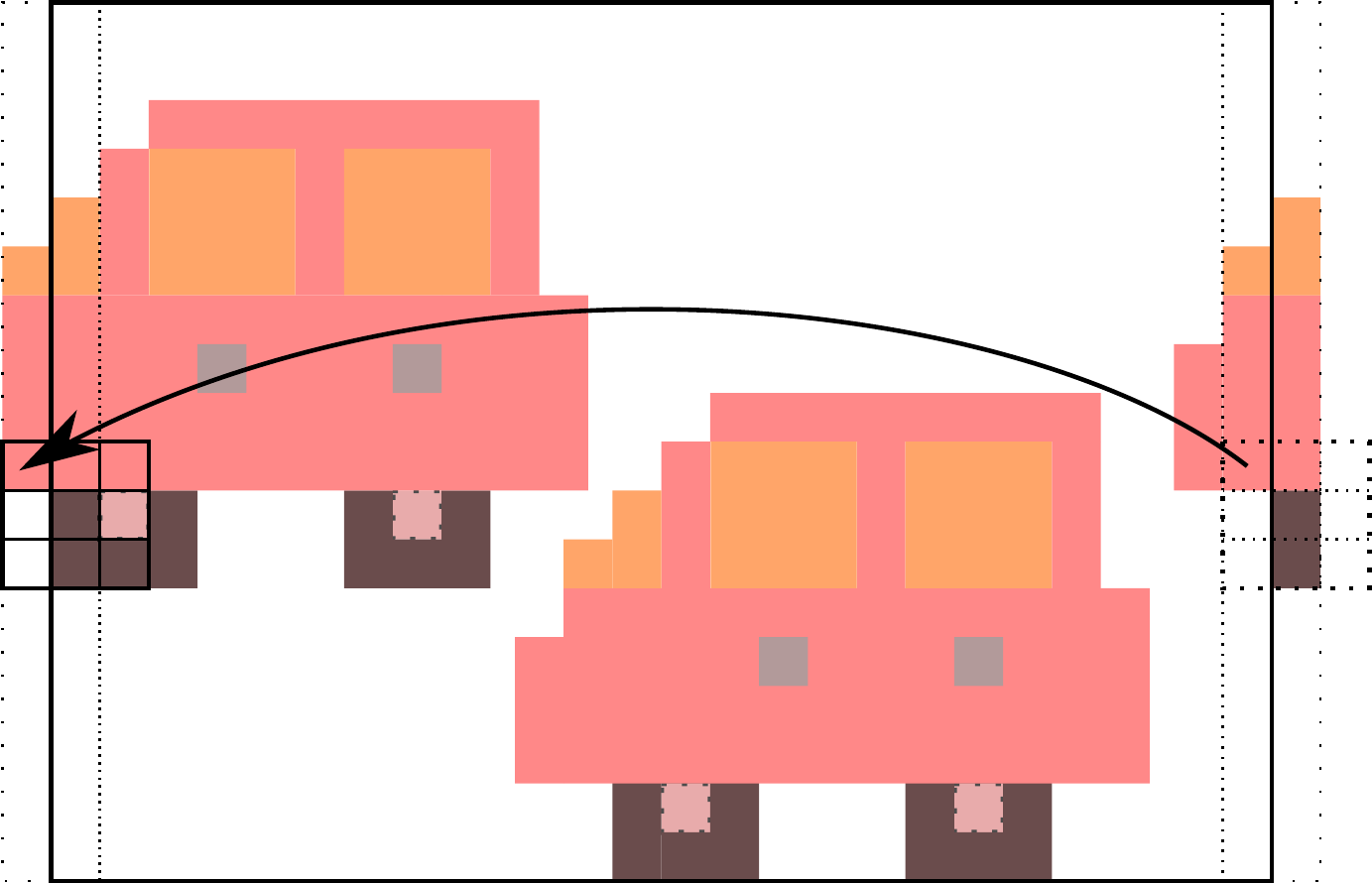}
}
    \caption{Convolutions are computed seamlessly across horizontal image boundaries using our proposed padding approach.}
\end{figure}

While the trained network can be used as is \cite{HuDeep360Pilot2017,DengObjectdetectionpanoramic2017} without any further modification, objects overlapping the left and right extremities of the equirectangular image would be split into two objects; one on the left, and one on the right (as depicted in Fig. \ref{fig:360folding}, bottom left). Moreover, information would not flow from one side of the image to the other side of the image --- at least in the early feature detection layers.
As a result, the deep architecture would \enquote{\emph{see}} those objects as if heavily occluded. Therefore, it is more difficult to detect objects overlapping the image boundary leading to decreased overall detection accuracy and recall.

A cropped equirectangular panorama can be folded into a 360$^\circ$ ring shown in Fig. \ref{fig:360folding} by stitching the left and right edges together. A 2D convolution on this ring is equivalent to padding the left and right side of the equirectangular image with respective pixels from the right and left side as if the image was tiled (as illustrated on Fig. \ref{fig:kernel3x3} for $3\times 3$ convolutions).
This horizontal ring-padding is hence used on all convolutional layers instead of the conventional zero-padding to eliminate these otherwise undesirable boundary effects.

For 3D detection, our proposed approach based on Faster R-CNN \cite{RenFasterRCNNRealTime2015} generates a sequence of detection proposals and subsequently pools a subregion around each proposal to further regress the final proposal location, class and 3D pose.
To adapt this operation, instead of clamping subregion coordinates by the equirectangular image extremities, we instead wrap horizontally the coordinates of each pixel within the box:

\begin{equation}
    u_{wrap} \equiv u\ (\text{mod}\ w)
\end{equation}

where $u$ is the horizontal coordinate of the pixel, $u_{wrap}$ the wrapped horizontal coordinate within the image and $w$ the image width.

As a result of this approach, we are hence able to hide the image boundary, as a result, enabling a true 360\deg processing of the equirectangular imagery.

\section{Evaluation}

We evaluate our approach both qualitatively on panoramic images from the crowd-sourced street-level imagery of Mapillary \cite{MapillaryMapillaryResearch} as well as quantitatively using synthetic data generated using the CARLA \cite{DosovitskiyCARLAOpenUrban2017} automotive environment simulator\footnote{for future comparison our code, models and evaluation data is publicly available at: \url{https://gdlg.github.io/panoramic}}.

\subsection{Qualitative Evaluation}

\begin{figure}
    \begin{tabular}{cc}
    \includegraphics[width=.5\textwidth]{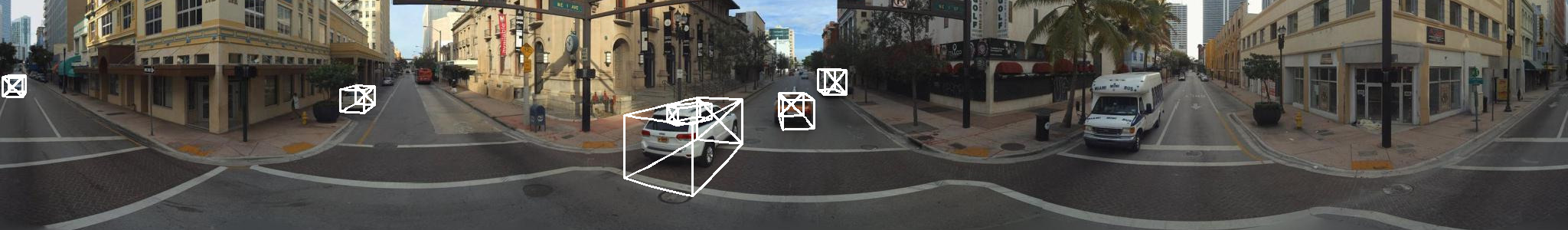} \includegraphics[width=.5\textwidth]{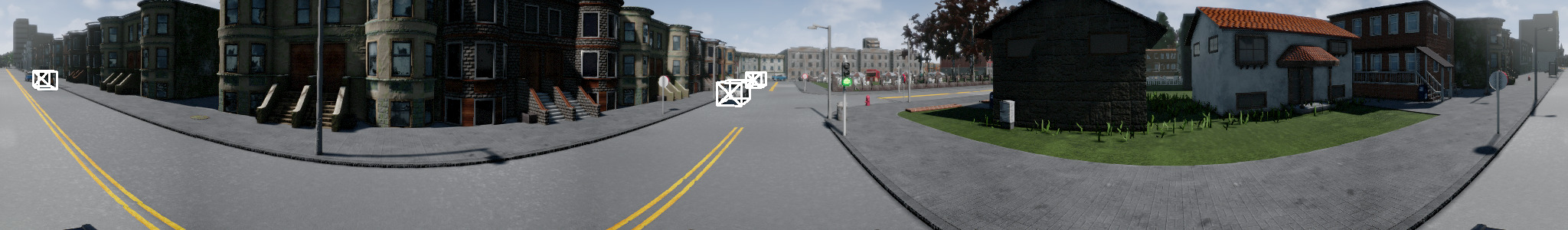} \\
    \includegraphics[width=.5\textwidth]{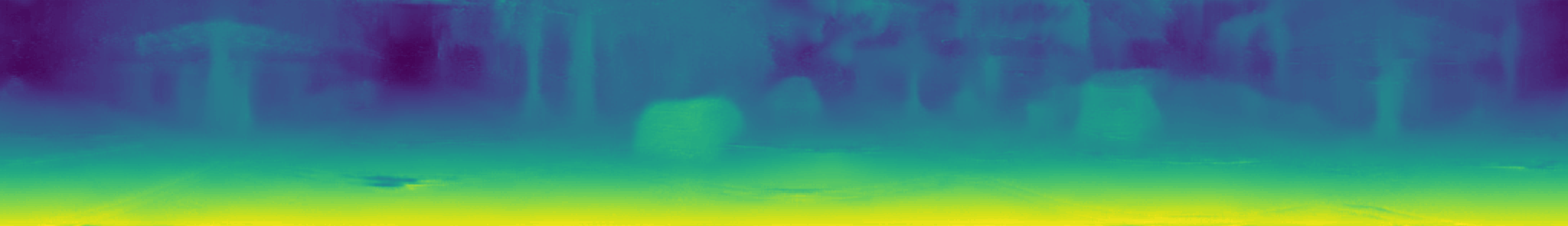} \includegraphics[width=.5\textwidth]{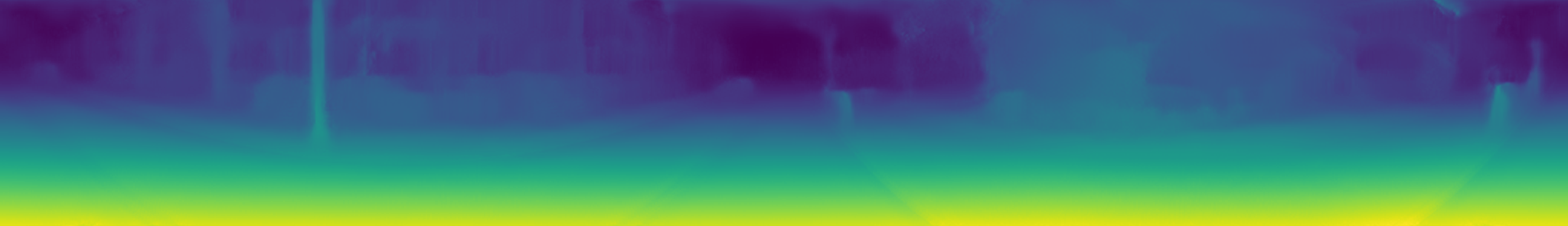} \\
    \includegraphics[width=.5\textwidth]{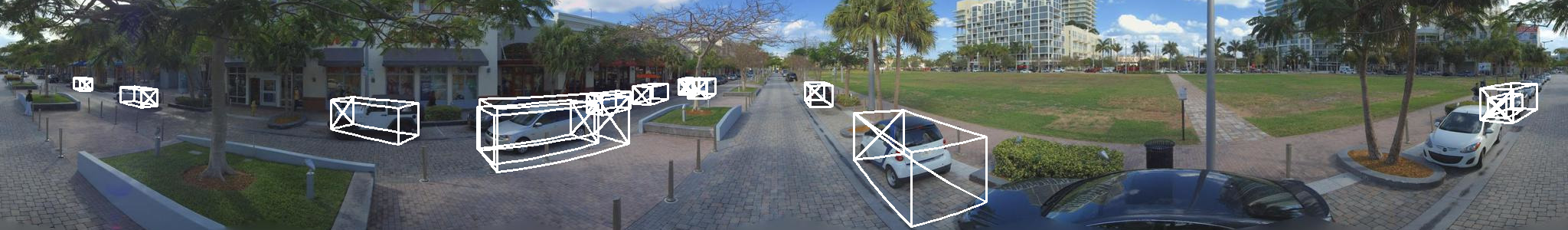} \includegraphics[width=.5\textwidth]{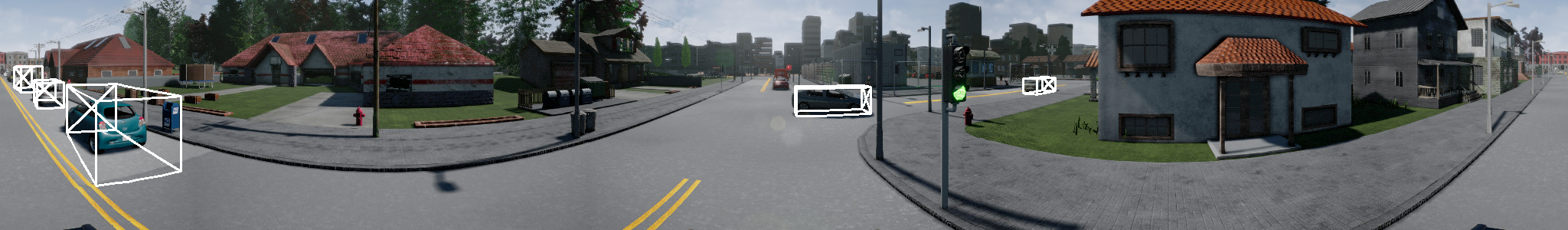} \\
    \includegraphics[width=.5\textwidth]{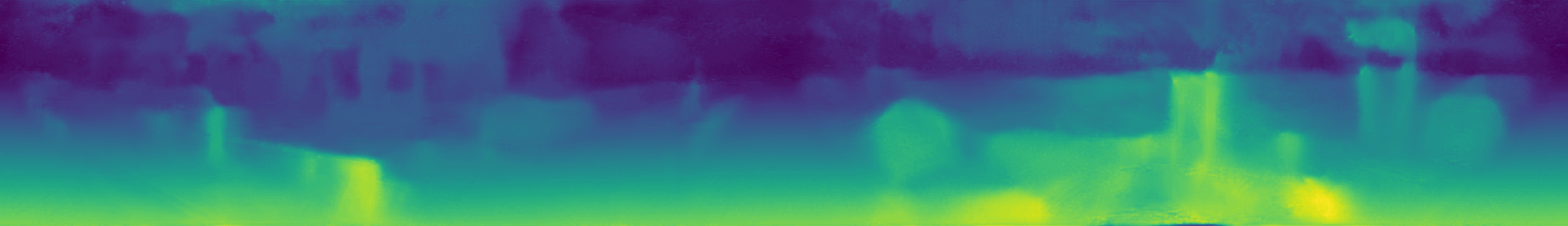} \includegraphics[width=.5\textwidth]{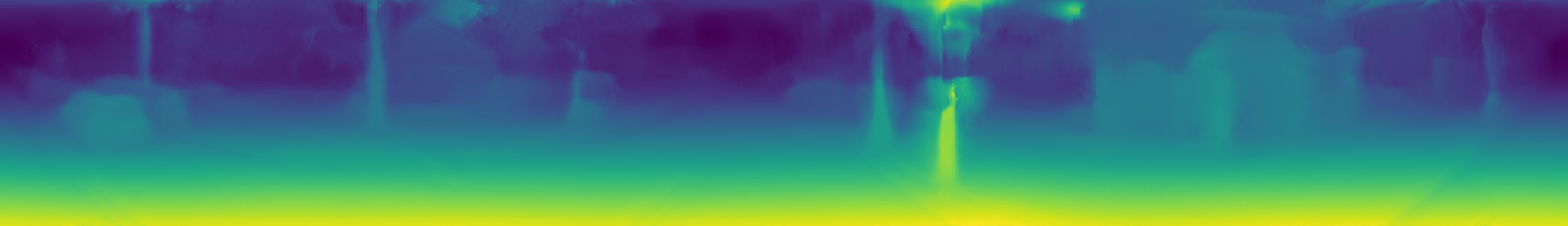} \\
    \includegraphics[width=.5\textwidth]{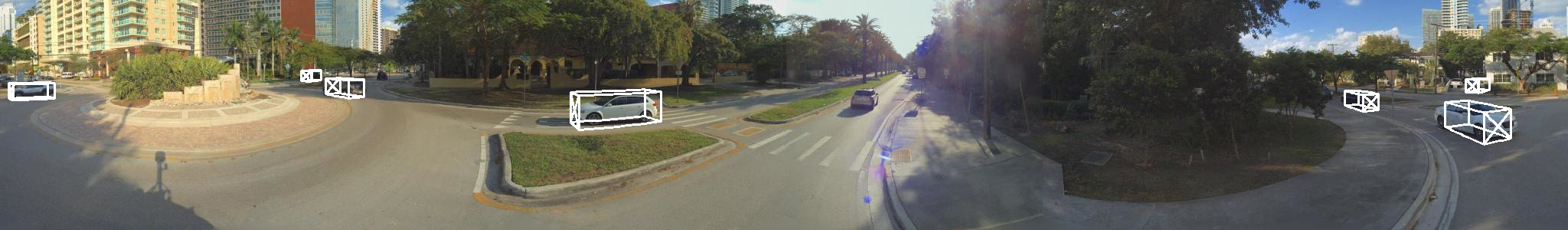} \includegraphics[width=.5\textwidth]{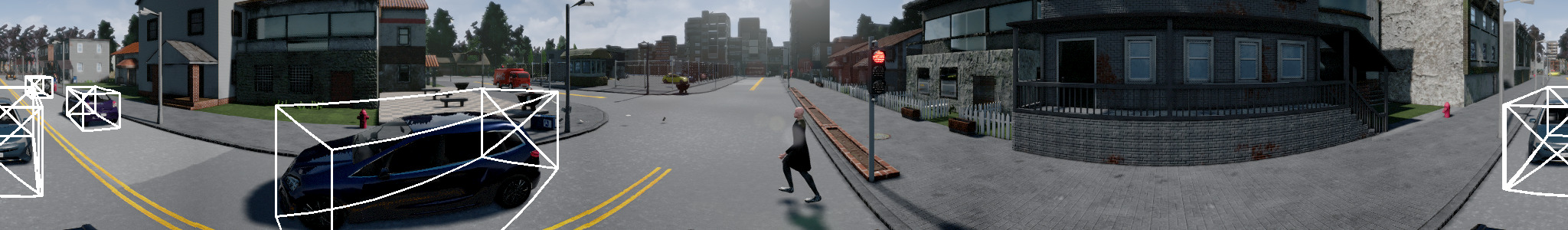} \\
    \includegraphics[width=.5\textwidth]{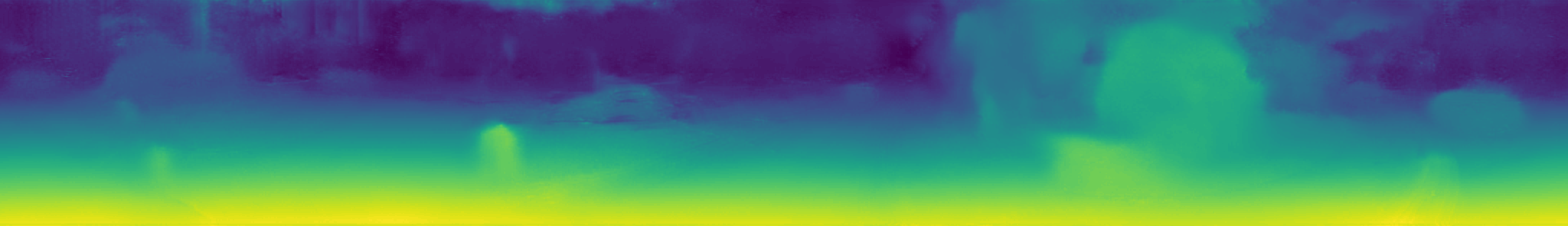} \includegraphics[width=.5\textwidth]{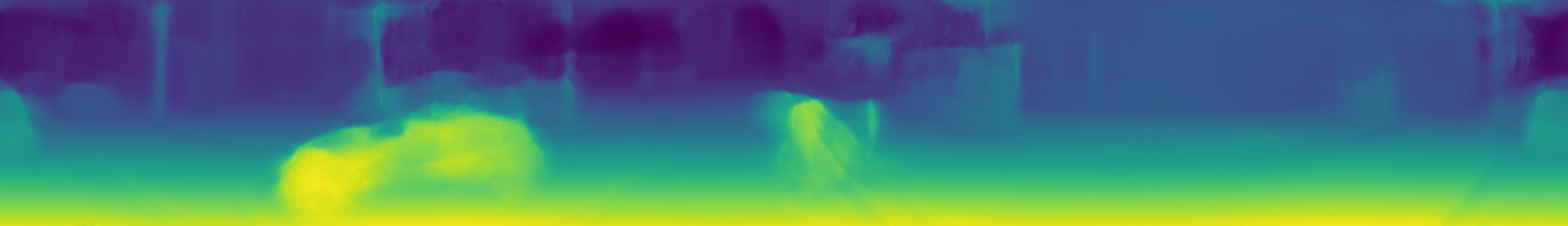} \\
    \includegraphics[width=.5\textwidth]{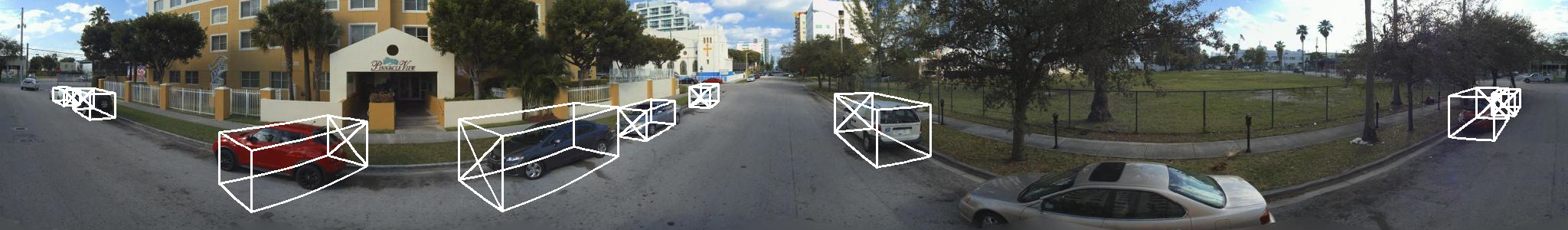} \includegraphics[width=.5\textwidth]{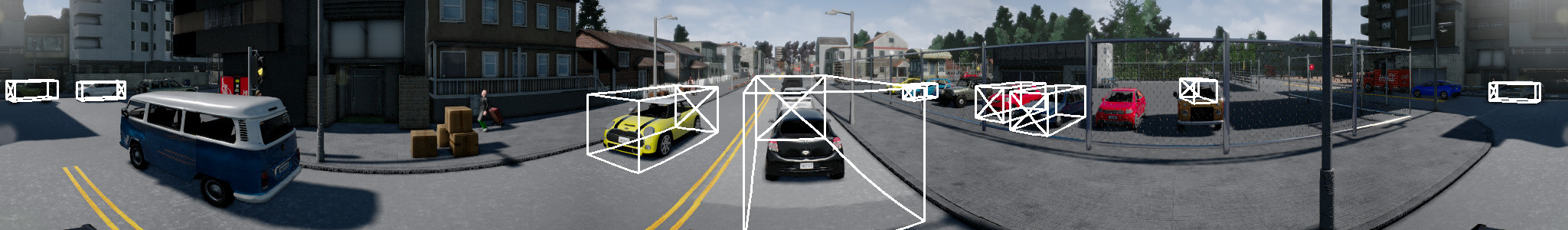} \\
    \includegraphics[width=.5\textwidth]{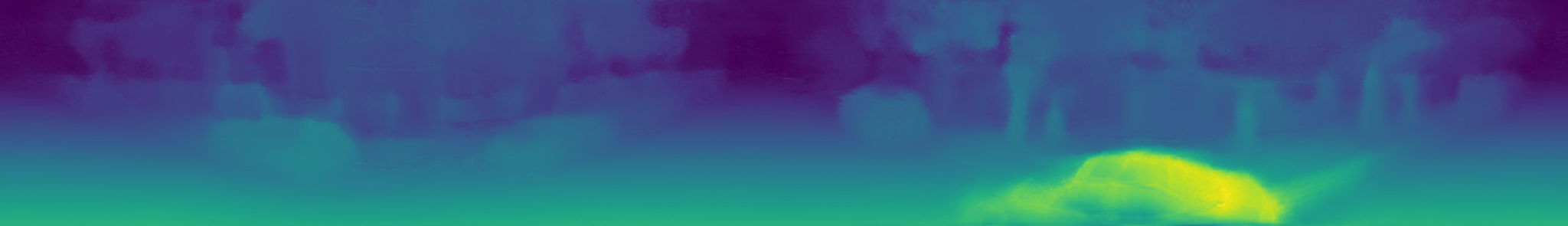} \includegraphics[width=.5\textwidth]{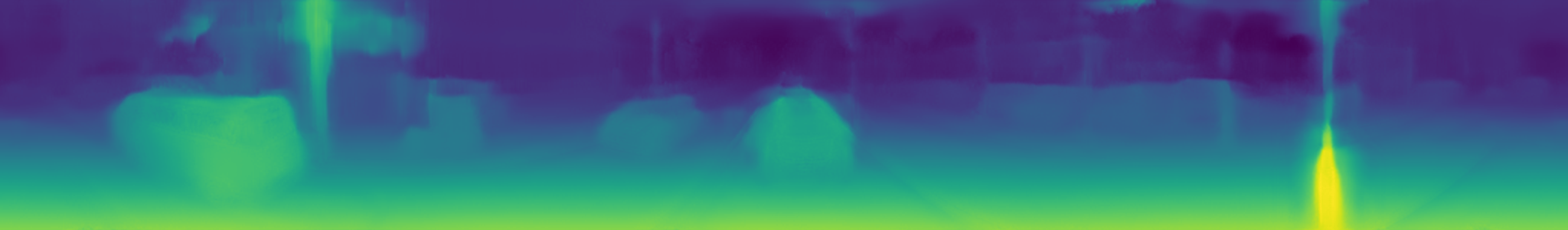} \\
    \end{tabular}
    \caption{\label{fig:results}Monocular depth recovery and 3D object detection with our approach. Left: Real-world images. Right: Synthetic images.}
    \vspace{-.2cm}
\end{figure}

\begin{figure}[t]
    \centering
    \includegraphics[height=1.5cm]{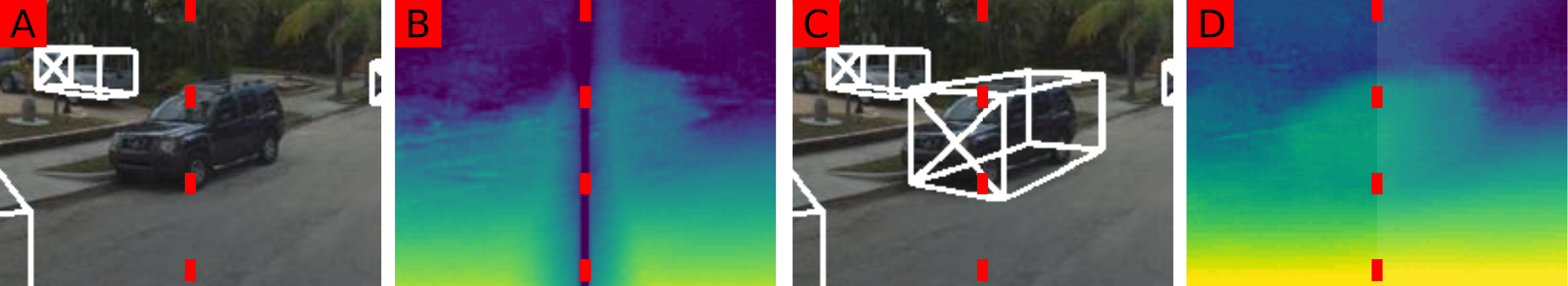}
    \caption{\label{fig:qual:ringpad}Right/left boundary effect. A,B: Zero-padding; C,D: Ring-padding.}
    \vspace{-.8cm}
\end{figure}

As discussed in Sec. \ref{sec:style_transfer}, we qualitatively evaluate our method using 30,000 panoramic images (Miami, USA) from the crowd-sourced street-level imagery of Mapillary \cite{MapillaryMapillaryResearch}. Fig. \ref{fig:results} shows our depth recovery and 3D object detection results on a selection of images of representative scenes from the data.
Ring-padding naturally enforces continuity across the right/left boundary; for instance, zero-padding can prevent detection of vehicles crossing the image boundary (Fig \ref{fig:qual:ringpad}A) whereas ring-padding seamlessly detects such vehicle (Fig. \ref{fig:qual:ringpad}C). Similarly zero-padding introduces depth discontinuities on the boundary (Fig \ref{fig:qual:ringpad}B) whereas ring-padding enforces depth continuity (Fig. \ref{fig:qual:ringpad}D).

The algorithm is able to successfully estimate the 3D pose of vehicles and recover scene depth. However the approach fails on vehicles which are too close to the camera, almost underneath the camera. Indeed, those view angles from above are not available in the narrow vertical field of view of the KITTI benchmark. Following the conventions of the KITTI dataset, any vehicles less than 25 pixels in image height were ignored during training. Due to the lower resolutions of the panoramic images, an average-size vehicle (about $2m$ height) with an apparent height of 25 pixels in KITTI is approximately at a distance of $56.6m$, whereas the same vehicle in a panoramic image will stand at $26m$. As a result, the range of the algorithm is reduced even though this is not a fundamental limitation of the approach itself. Rather, we expect this maximum distance to be increased as the resolution of the panoramic imagery is increased.

Further results are available in the supplementary video\footnotemark[2].

\subsection{Quantitative Evaluation Methodology}

Due to the lack of available annotated automotive panoramic imagery dataset, we evaluate our algorithm on synthetic data generated using the CARLA automotive environment simulator \cite{DosovitskiyCARLAOpenUrban2017} adapted for panoramic imagery rendering using the same format as our qualitative dataset. Due to lack of variety, our dataset based on CARLA is not suitable for training purposes, while it is suitable for cross-dataset validation.
Following KITTI conventions, we filtered out vehicles less than 25 pixels in height from our detection results. 

\begin{figure}
    \centering
    \subfigure[\label{fig:2d_detection_results}Object detection results]{
    \resizebox{.45\textwidth}{!}{\large\input{detections/car_detection}}
}
    \subfigure[3D Intersection over Union]{
    \resizebox{.45\textwidth}{!}{\large\input{detections/car_iou3d}}
}
    \caption{Object detection results}
\end{figure}
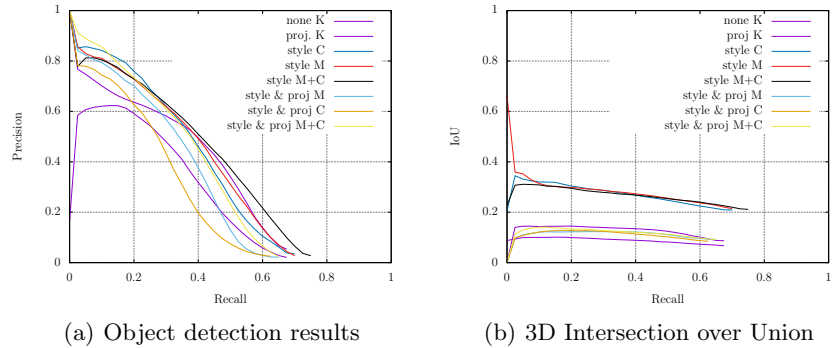

Table \ref{tab:detection} shows the \emph{mean average precison} (mAP) using an \emph{intersection over union} (IoU) of 0.5 across variations of our algorithm on 8,000 images. Overall, the projection transformation during training impairs the results by about 10\% points. Our best results come from the combined style-transferred training dataset consisting of both Mapillary and CARLA (4\% points increased compared to original) whilst training on the CARLA-adapted dataset alone increases the performance by 2\% points.  
This is due to the simplistic rendering and lack of variety of the synthetic dataset which impairs the style transfer. As a result, the CARLA-adapted dataset significantly boosts the accuracy for very low recall; however, it also reduces the recall ability of the network (Fig. \ref{fig:2d_detection_results}).
The model trained on the CARLA-adapted dataset achieves a mAP of 0.82 on our evaluation set of the adapted images but only 0.35 on the actual CARLA dataset which shows that the style transfer is somewhat limited.
Qualitatively, style transfer toward the Mapillary dataset, which is of similar scene complexity to KITTI, is significantly better than CARLA. By contrast, the combined dataset is able to outperform on both metrics (Fig. \ref{fig:2d_detection_results}). 

The monocular depth estimation results are shown in Table \ref{tab:detection} for 200 images (for distances $<50m$). Similar to our detection result, using CARLA-adapted imagery impairs the performance. Using projection transformation, we see an increase of about 2.5\% points in accuracy. Overall, those differences are smaller than those on object detection across the different transformations (Table \ref{tab:detection}).

\begin{table}[t]
    \centering
    \begin{threeparttable}
    \begin{tabular}{lc|c|llll|c|}
        & & Detection\tnote{a}$\ $\ & \multicolumn{4}{c|}{Depth Error Metrics\tnote{b}} & \multicolumn{1}{c|}{Depth Acc.\tnote{a}$\ $} \\
        Transformation & Dataset & mAP & Abs. rel.   & Sq. rel.   & RMSE      & RMSE log & $\delta < 1.25$ \\
        \hline \hline
        none & K & 0.336 & 0.247 & 7.652 & 3.484 & 0.465 & 0.697\\
        proj. & K & 0.244 & 0.251 & 7.381 & 3.451 & 0.445 & 0.732 \\
        style & C & 0.355 & 0.262 & 7.668 & 3.601 & 0.480 & 0.686 \\
        style & M & 0.359 & 0.257 & 7.937 & 3.634 & 0.474 & 0.682 \\
        style & M+C & 0.378  & 0.230 & 6.338 & 3.619 & 0.474 & 0.679 \\
        style \& proj. & C & 0.259 & 0.292 & 9.649 & 3.660 & 0.469 & 0.723 \\
        style \& proj. & M & 0.308 & 0.300 & 10.467 & 3.798 & 0.473 & 0.719 \\
        style \& proj. & M+C & 0.344 & 0.231 & 6.377 & 3.598 & 0.463 & 0.716 \\
    \end{tabular}
        \vspace{5pt}
        \begin{tablenotes}
        \item{a} Higher, better \item{b} Lower, better
        \end{tablenotes}
    \end{threeparttable}
        \vspace{5pt}
    \caption{\label{tab:detection} 3D Object detection (mAP) results; and depth recovery results using metrics defined by \cite{EigenDepthMapPrediction2014}. Training dataset: C: CARLA, M: Mapillary, K: KITTI}
\end{table}

From our results, we can clearly see that we have identified a new and challenging problem within the automotive visual sensing space (Table \ref{tab:detection}) when compared to the rectilinear performance of contemporary benchmarks \cite{GeigerAreweready2012,GeigerVisionmeetsrobotics2013}.

\section{Conclusion}

We have adapted existing deep architectures and training datasets, proven on forward-facing rectilinear camera imagery, to perform on panoramic images. The approach is based on domain adaptation using geometrical and style transforms and novel updates to training loss to accommodate panoramic imagery.
Our approach is able to recover the monocular depth and the full 3D pose of vehicles.

We have identified panoramic imagery has a new set of challenging problems in automotive visual sensing and provide the first performance benchmark for the use of these techniques on 360\deg panoramic imagery, with a supporting dataset, hence acting as a key driver for future research on this topic.

\bibliographystyle{splncs04}
\bibliography{biblio,amir}

\end{document}

%% file: detections/car_detection.tex
\begingroup
  \makeatletter
  \providecommand\color[2][]{%
    \GenericError{(gnuplot) \space\space\space\@spaces}{%
      Package color not loaded in conjunction with
      terminal option `colourtext'%
    }{See the gnuplot documentation for explanation.%
    }{Either use 'blacktext' in gnuplot or load the package
      color.sty in LaTeX.}%
    \renewcommand\color[2][]{}%
  }%
  \providecommand\includegraphics[2][]{%
    \GenericError{(gnuplot) \space\space\space\@spaces}{%
      Package graphicx or graphics not loaded%
    }{See the gnuplot documentation for explanation.%
    }{The gnuplot epslatex terminal needs graphicx.sty or graphics.sty.}%
    \renewcommand\includegraphics[2][]{}%
  }%
  \providecommand\rotatebox[2]{#2}%
  \@ifundefined{ifGPcolor}{%
    \newif\ifGPcolor
    \GPcolorfalse
  }{}%
  \@ifundefined{ifGPblacktext}{%
    \newif\ifGPblacktext
    \GPblacktexttrue
  }{}%
  \let\gplgaddtomacro\g@addto@macro
  \gdef\gplbacktext{}%
  \gdef\gplfronttext{}%
  \makeatother
  \ifGPblacktext
    \def\colorrgb#1{}%
    \def\colorgray#1{}%
  \else
    \ifGPcolor
      \def\colorrgb#1{\color[rgb]{#1}}%
      \def\colorgray#1{\color[gray]{#1}}%
      \expandafter\def\csname LTw\endcsname{\color{white}}%
      \expandafter\def\csname LTb\endcsname{\color{black}}%
      \expandafter\def\csname LTa\endcsname{\color{black}}%
      \expandafter\def\csname LT0\endcsname{\color[rgb]{1,0,0}}%
      \expandafter\def\csname LT1\endcsname{\color[rgb]{0,1,0}}%
      \expandafter\def\csname LT2\endcsname{\color[rgb]{0,0,1}}%
      \expandafter\def\csname LT3\endcsname{\color[rgb]{1,0,1}}%
      \expandafter\def\csname LT4\endcsname{\color[rgb]{0,1,1}}%
      \expandafter\def\csname LT5\endcsname{\color[rgb]{1,1,0}}%
      \expandafter\def\csname LT6\endcsname{\color[rgb]{0,0,0}}%
      \expandafter\def\csname LT7\endcsname{\color[rgb]{1,0.3,0}}%
      \expandafter\def\csname LT8\endcsname{\color[rgb]{0.5,0.5,0.5}}%
    \else
      \def\colorrgb#1{\color{black}}%
      \def\colorgray#1{\color[gray]{#1}}%
      \expandafter\def\csname LTw\endcsname{\color{white}}%
      \expandafter\def\csname LTb\endcsname{\color{black}}%
      \expandafter\def\csname LTa\endcsname{\color{black}}%
      \expandafter\def\csname LT0\endcsname{\color{black}}%
      \expandafter\def\csname LT1\endcsname{\color{black}}%
      \expandafter\def\csname LT2\endcsname{\color{black}}%
      \expandafter\def\csname LT3\endcsname{\color{black}}%
      \expandafter\def\csname LT4\endcsname{\color{black}}%
      \expandafter\def\csname LT5\endcsname{\color{black}}%
      \expandafter\def\csname LT6\endcsname{\color{black}}%
      \expandafter\def\csname LT7\endcsname{\color{black}}%
      \expandafter\def\csname LT8\endcsname{\color{black}}%
    \fi
  \fi
    \setlength{\unitlength}{0.0500bp}%
    \ifx\gptboxheight\undefined%
      \newlength{\gptboxheight}%
      \newlength{\gptboxwidth}%
      \newsavebox{\gptboxtext}%
    \fi%
    \setlength{\fboxrule}{0.5pt}%
    \setlength{\fboxsep}{1pt}%
\begin{picture}(7920.00,6048.00)%
    \gplgaddtomacro\gplbacktext{%
      \csname LTb\endcsname
      \put(1062,921){\makebox(0,0)[r]{\strut{}$0$}}%
      \csname LTb\endcsname
      \put(1062,1889){\makebox(0,0)[r]{\strut{}$0.2$}}%
      \csname LTb\endcsname
      \put(1062,2856){\makebox(0,0)[r]{\strut{}$0.4$}}%
      \csname LTb\endcsname
      \put(1062,3824){\makebox(0,0)[r]{\strut{}$0.6$}}%
      \csname LTb\endcsname
      \put(1062,4791){\makebox(0,0)[r]{\strut{}$0.8$}}%
      \csname LTb\endcsname
      \put(1062,5759){\makebox(0,0)[r]{\strut{}$1$}}%
      \csname LTb\endcsname
      \put(1234,633){\makebox(0,0){\strut{}$0$}}%
      \csname LTb\endcsname
      \put(2468,633){\makebox(0,0){\strut{}$0.2$}}%
      \csname LTb\endcsname
      \put(3701,633){\makebox(0,0){\strut{}$0.4$}}%
      \csname LTb\endcsname
      \put(4935,633){\makebox(0,0){\strut{}$0.6$}}%
      \csname LTb\endcsname
      \put(6168,633){\makebox(0,0){\strut{}$0.8$}}%
      \csname LTb\endcsname
      \put(7402,633){\makebox(0,0){\strut{}$1$}}%
    }%
    \gplgaddtomacro\gplfronttext{%
      \csname LTb\endcsname
      \put(258,3340){\rotatebox{-270}{\makebox(0,0){\strut{}Precision}}}%
      \put(4318,201){\makebox(0,0){\strut{}Recall}}%
      \csname LTb\endcsname
      \put(6135,5552){\makebox(0,0)[r]{\strut{}none K}}%
      \csname LTb\endcsname
      \put(6135,5264){\makebox(0,0)[r]{\strut{}proj. K}}%
      \csname LTb\endcsname
      \put(6135,4976){\makebox(0,0)[r]{\strut{}style C}}%
      \csname LTb\endcsname
      \put(6135,4688){\makebox(0,0)[r]{\strut{}style M}}%
      \csname LTb\endcsname
      \put(6135,4400){\makebox(0,0)[r]{\strut{}style M+C}}%
      \csname LTb\endcsname
      \put(6135,4112){\makebox(0,0)[r]{\strut{}style \& proj M}}%
      \csname LTb\endcsname
      \put(6135,3824){\makebox(0,0)[r]{\strut{}style \& proj C}}%
      \csname LTb\endcsname
      \put(6135,3536){\makebox(0,0)[r]{\strut{}style \& proj M+C}}%
    }%
    \gplbacktext
    \put(0,0){\includegraphics{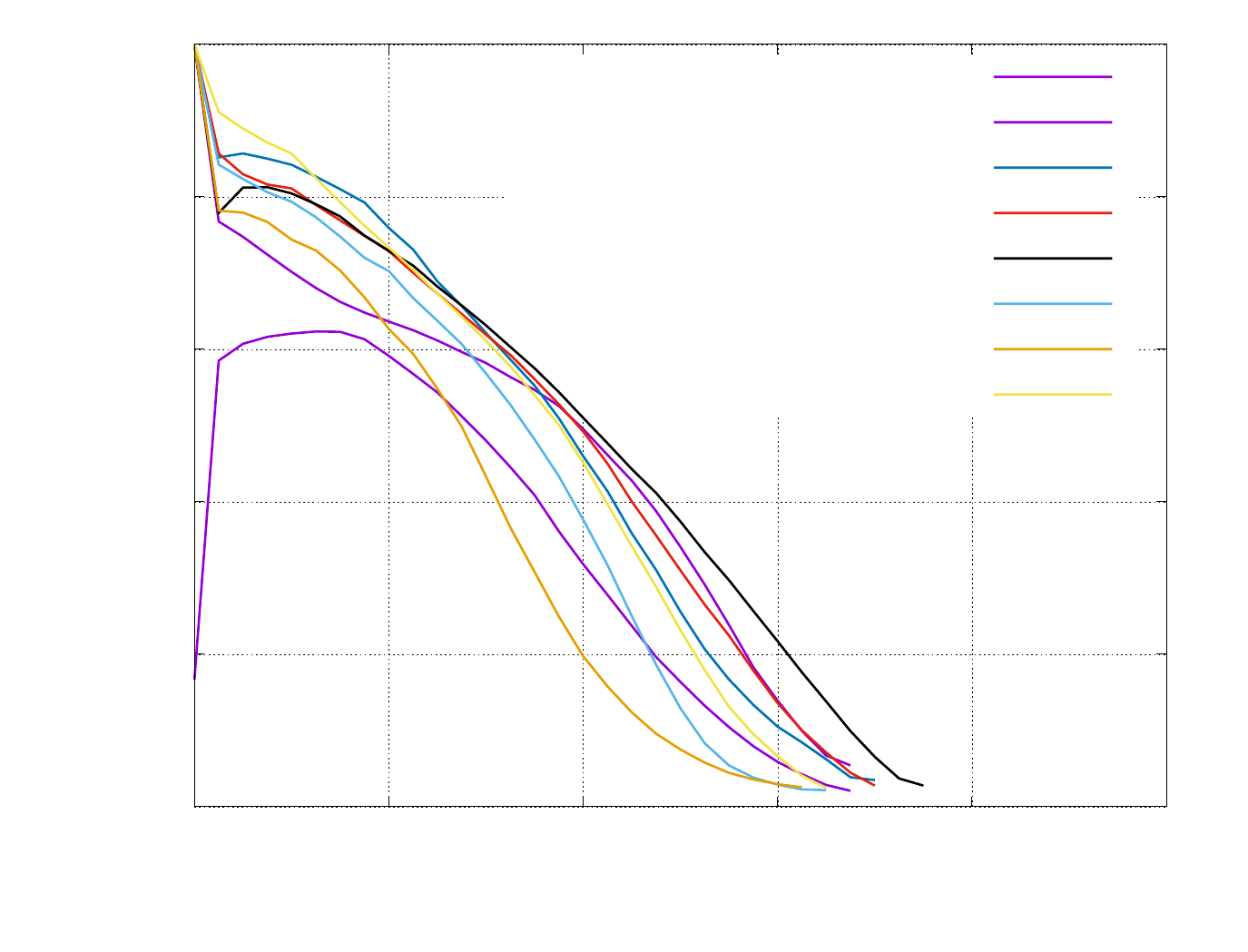}}%
    \gplfronttext
  \end{picture}%
\endgroup

%% file: detections/car_iou3d.tex
\begingroup
  \makeatletter
  \providecommand\color[2][]{%
    \GenericError{(gnuplot) \space\space\space\@spaces}{%
      Package color not loaded in conjunction with
      terminal option `colourtext'%
    }{See the gnuplot documentation for explanation.%
    }{Either use 'blacktext' in gnuplot or load the package
      color.sty in LaTeX.}%
    \renewcommand\color[2][]{}%
  }%
  \providecommand\includegraphics[2][]{%
    \GenericError{(gnuplot) \space\space\space\@spaces}{%
      Package graphicx or graphics not loaded%
    }{See the gnuplot documentation for explanation.%
    }{The gnuplot epslatex terminal needs graphicx.sty or graphics.sty.}%
    \renewcommand\includegraphics[2][]{}%
  }%
  \providecommand\rotatebox[2]{#2}%
  \@ifundefined{ifGPcolor}{%
    \newif\ifGPcolor
    \GPcolorfalse
  }{}%
  \@ifundefined{ifGPblacktext}{%
    \newif\ifGPblacktext
    \GPblacktexttrue
  }{}%
  \let\gplgaddtomacro\g@addto@macro
  \gdef\gplbacktext{}%
  \gdef\gplfronttext{}%
  \makeatother
  \ifGPblacktext
    \def\colorrgb#1{}%
    \def\colorgray#1{}%
  \else
    \ifGPcolor
      \def\colorrgb#1{\color[rgb]{#1}}%
      \def\colorgray#1{\color[gray]{#1}}%
      \expandafter\def\csname LTw\endcsname{\color{white}}%
      \expandafter\def\csname LTb\endcsname{\color{black}}%
      \expandafter\def\csname LTa\endcsname{\color{black}}%
      \expandafter\def\csname LT0\endcsname{\color[rgb]{1,0,0}}%
      \expandafter\def\csname LT1\endcsname{\color[rgb]{0,1,0}}%
      \expandafter\def\csname LT2\endcsname{\color[rgb]{0,0,1}}%
      \expandafter\def\csname LT3\endcsname{\color[rgb]{1,0,1}}%
      \expandafter\def\csname LT4\endcsname{\color[rgb]{0,1,1}}%
      \expandafter\def\csname LT5\endcsname{\color[rgb]{1,1,0}}%
      \expandafter\def\csname LT6\endcsname{\color[rgb]{0,0,0}}%
      \expandafter\def\csname LT7\endcsname{\color[rgb]{1,0.3,0}}%
      \expandafter\def\csname LT8\endcsname{\color[rgb]{0.5,0.5,0.5}}%
    \else
      \def\colorrgb#1{\color{black}}%
      \def\colorgray#1{\color[gray]{#1}}%
      \expandafter\def\csname LTw\endcsname{\color{white}}%
      \expandafter\def\csname LTb\endcsname{\color{black}}%
      \expandafter\def\csname LTa\endcsname{\color{black}}%
      \expandafter\def\csname LT0\endcsname{\color{black}}%
      \expandafter\def\csname LT1\endcsname{\color{black}}%
      \expandafter\def\csname LT2\endcsname{\color{black}}%
      \expandafter\def\csname LT3\endcsname{\color{black}}%
      \expandafter\def\csname LT4\endcsname{\color{black}}%
      \expandafter\def\csname LT5\endcsname{\color{black}}%
      \expandafter\def\csname LT6\endcsname{\color{black}}%
      \expandafter\def\csname LT7\endcsname{\color{black}}%
      \expandafter\def\csname LT8\endcsname{\color{black}}%
    \fi
  \fi
    \setlength{\unitlength}{0.0500bp}%
    \ifx\gptboxheight\undefined%
      \newlength{\gptboxheight}%
      \newlength{\gptboxwidth}%
      \newsavebox{\gptboxtext}%
    \fi%
    \setlength{\fboxrule}{0.5pt}%
    \setlength{\fboxsep}{1pt}%
\begin{picture}(7920.00,6048.00)%
    \gplgaddtomacro\gplbacktext{%
      \csname LTb\endcsname
      \put(1062,921){\makebox(0,0)[r]{\strut{}$0$}}%
      \csname LTb\endcsname
      \put(1062,1889){\makebox(0,0)[r]{\strut{}$0.2$}}%
      \csname LTb\endcsname
      \put(1062,2856){\makebox(0,0)[r]{\strut{}$0.4$}}%
      \csname LTb\endcsname
      \put(1062,3824){\makebox(0,0)[r]{\strut{}$0.6$}}%
      \csname LTb\endcsname
      \put(1062,4791){\makebox(0,0)[r]{\strut{}$0.8$}}%
      \csname LTb\endcsname
      \put(1062,5759){\makebox(0,0)[r]{\strut{}$1$}}%
      \csname LTb\endcsname
      \put(1234,633){\makebox(0,0){\strut{}$0$}}%
      \csname LTb\endcsname
      \put(2468,633){\makebox(0,0){\strut{}$0.2$}}%
      \csname LTb\endcsname
      \put(3701,633){\makebox(0,0){\strut{}$0.4$}}%
      \csname LTb\endcsname
      \put(4935,633){\makebox(0,0){\strut{}$0.6$}}%
      \csname LTb\endcsname
      \put(6168,633){\makebox(0,0){\strut{}$0.8$}}%
      \csname LTb\endcsname
      \put(7402,633){\makebox(0,0){\strut{}$1$}}%
    }%
    \gplgaddtomacro\gplfronttext{%
      \csname LTb\endcsname
      \put(258,3340){\rotatebox{-270}{\makebox(0,0){\strut{}IoU}}}%
      \put(4318,201){\makebox(0,0){\strut{}Recall}}%
      \csname LTb\endcsname
      \put(6135,5552){\makebox(0,0)[r]{\strut{}none K}}%
      \csname LTb\endcsname
      \put(6135,5264){\makebox(0,0)[r]{\strut{}proj K}}%
      \csname LTb\endcsname
      \put(6135,4976){\makebox(0,0)[r]{\strut{}style C}}%
      \csname LTb\endcsname
      \put(6135,4688){\makebox(0,0)[r]{\strut{}style M}}%
      \csname LTb\endcsname
      \put(6135,4400){\makebox(0,0)[r]{\strut{}style M+C}}%
      \csname LTb\endcsname
      \put(6135,4112){\makebox(0,0)[r]{\strut{}style \& proj M}}%
      \csname LTb\endcsname
      \put(6135,3824){\makebox(0,0)[r]{\strut{}style \& proj C}}%
      \csname LTb\endcsname
      \put(6135,3536){\makebox(0,0)[r]{\strut{}style \& proj M+C}}%
    }%
    \gplbacktext
    \put(0,0){\includegraphics{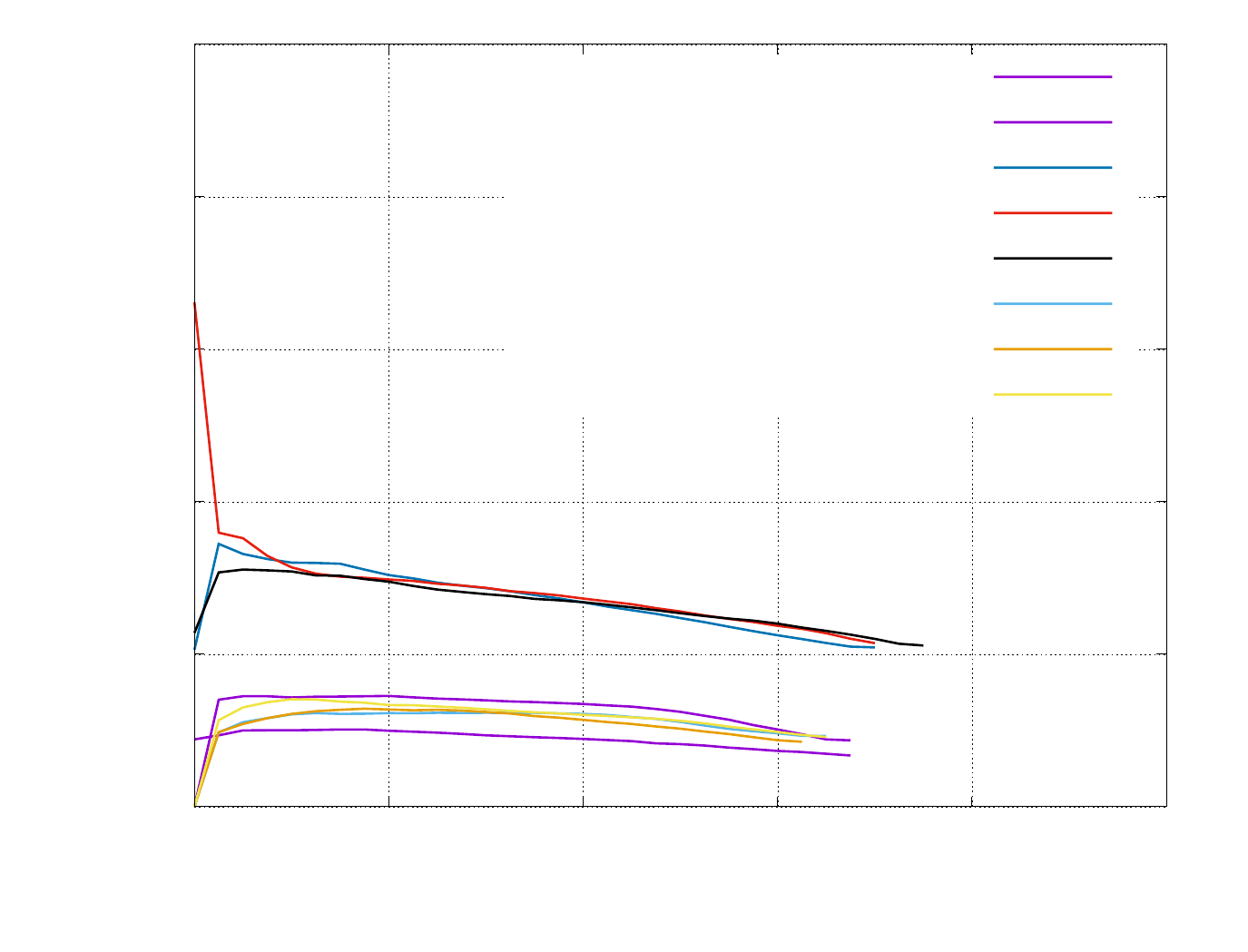}}%
    \gplfronttext
  \end{picture}%
\endgroup